\documentclass[sigconf]{acmart}
\acmSubmissionID{542}

\usepackage{booktabs} 

\usepackage{bm}

\usepackage[ruled]{algorithm2e} 

\usepackage{algpseudocode}
\SetAlFnt{\small}
\SetAlCapFnt{\small}
\SetAlCapNameFnt{\small}
\SetAlCapHSkip{0pt}

\acmJournal{TOG}

\copyrightyear{2023} 
\acmYear{2023} 
\setcopyright{acmlicensed}\acmConference[SIGGRAPH '23 Conference Proceedings]{Special Interest Group on Computer Graphics and Interactive Techniques Conference Conference Proceedings}{August 6--10, 2023}{Los Angeles, CA, USA}
\acmBooktitle{Special Interest Group on Computer Graphics and Interactive Techniques Conference Conference Proceedings (SIGGRAPH '23 Conference Proceedings), August 6--10, 2023, Los Angeles, CA, USA}
\acmPrice{15.00}
\acmDOI{10.1145/3588432.3591528}
\acmISBN{979-8-4007-0159-7/23/08}

\begin{document}
\title{Synthesizing Dexterous Nonprehensile Pregrasp for Ungraspable Objects}

\author{Sirui Chen}
\affiliation{%
 \institution{University of Hong Kong}
 \country{HKSAR, China}}
\email{ericcsr@connect.hku.hk}
\author{Albert Wu}
\affiliation{%
 \institution{Stanford University}
 \state{California}
 \country{USA}
}
\email{amhwu@stanford.edu}
\author{C. Karen Liu}
\affiliation{%
\institution{Stanford University}
\state{California}
\country{USA}}
\email{karenliu@cs.stanford.edu}

\begin{teaserfigure}
  \includegraphics[width=\textwidth]{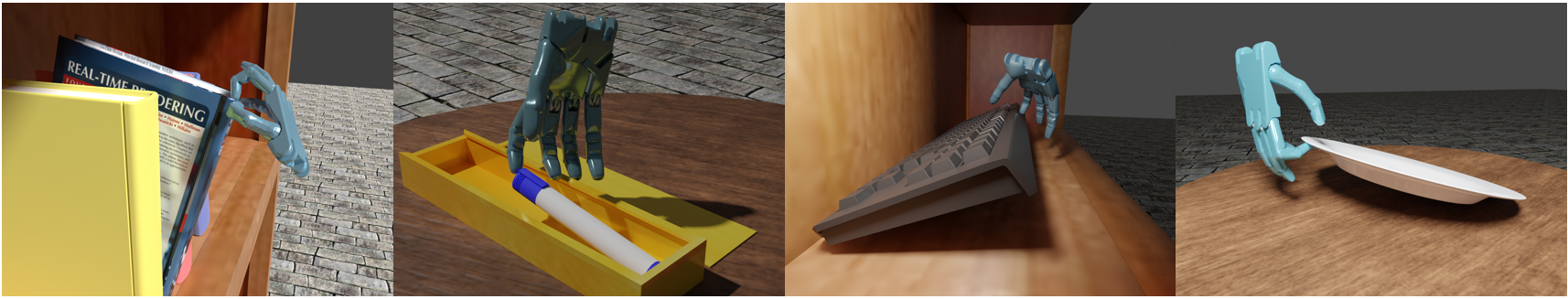}
  \caption{We propose a method to synthesize physically realistic nonprehensile pregrasp motions. Our method automatically discovers various strategies to leverage contacts with the surrounding environment and the hand }
  \Description{This is the teaser figure for the article.}
  \label{fig:teaser}
\end{teaserfigure}
\begin{abstract}

Daily objects embedded in a contextual environment are often ungraspable initially. Whether it is a book sandwiched by other books on a fully packed bookshelf or a piece of paper lying flat on the desk, a series of nonprehensile pregrasp maneuvers is required to manipulate the object into a graspable state. Humans are proficient at utilizing environmental contacts to achieve manipulation tasks that are otherwise impossible, but synthesizing such nonprehensile pregrasp behaviors is challenging to existing methods. We present a novel method that combines graph search, optimal control, and a learning-based objective function to synthesize physically realistic and diverse nonprehensile pre-grasp motions that leverage the external contacts. Since the ``graspability'' of an object in context with its surrounding is difficult to define, we utilize a dataset of dexterous grasps to learn a metric which implicitly takes into account the exposed surface of the object and the finger tip locations. Our method can efficiently discover hand and object trajectories that are certified to be physically feasible by the simulation and kinematically achievable by the dexterous hand. We evaluate our method on eight challenging scenarios where nonprehensile pre-grasps are required to succeed. We also show that our method can be applied to unseen objects different from those in the training dataset. Finally, we report quantitative analyses on generalization and robustness of our method, as well as an ablation study.

\end{abstract}

%
%


%
%


\maketitle

\section{Introduction}
Manipulating objects with a dexterous multi-fingered hand is a key human ability. In particular, humans are proficient at leveraging environmental contacts to perform tasks that are otherwise impossible, utilizing \textit{nonprehensile manipulation}, a strategy to move objects without establishing a firm grasp first. For instance, when removing a book from a densely packed bookshelf, one would pivot the book outwards with one finger while keeping the book stabilized with lateral environment contacts. Once sufficient area of the book is exposed, the book can be picked up with a simple grasp (Figure ~\ref{fig:teaser}). This type of non-prehensile ``pregrasps'' is an essential skill for operating in an ecological human living space in which objects are often initially ungraspable due to occlusions by other surrounding objects.

Replicating human-like nonprehensile manipulation is challenging to existing motion planning methods. The inclusion of environmental contacts leads to combinatorial complexity in the number of possible contact configurations, which is intractable to even state-of-the-art motion planners. Moreover, it is difficult to define heuristics for the purpose of trajectory optimization. Consequently, many papers in both the dexterous grasp generation literature (e.g.,~\cite{grasptta, graspfield, grabnet}) and the motion generation literature (e.g.,~\cite{manipnet}) involve motion generation through learning-based methods. Nevertheless, the rich contact constraints in manipulation makes it difficult for learning methods to produce physically realistic motions with no interpenetrating or telekinetic interactions. The tradeoff between computational efficiency and physical realism remains a major hurdle in synthesizing manipulation motion.


In this paper, we propose a method to synthesize physically realistic nonprehensile pregrasp motions that make ungraspable objects in cluttered environments graspable. In particular, this is achieved through leveraging contacts with the surrounding environment and the hand. Our key observation is that, for any object placed in the environment, \emph{at most two} finger contacts are necessary for nonprehensile pregrasp since there exist at least one external contact point which collectively achieves wrench closure with the two finger contacts. This significantly reduces the search space of our motion synthesis problem, and allows us to formulate pregrasp planning as a reasonably sized mixed-integer optimization problem. Under this observation, our method solves for sequences of at most two contact points on the object by formulating a combination of discrete graph search and trajectory optimization.

Defining an exact metric for  “graspability” during the pregrasp phase is not trivial. The obvious goal in this phase is to expose the surface of the object, but it has to be done in such a way that the finger contacts during the pregrasp phase can be fluidly transitioned to a firm grasp later. As such, the graspability we would like to maximize depends on the current state of the object, the environment, the fingers that have established contact points and the fingers that are still free of contact. To define a generalizable metric without relying on heuristics, we learn a general metric of “graspability” through a dataset of dexterous grasps in various scenarios. Using this metric, we perform particle-based trajectory optimization in a physics-based simulator. This allows our method to efficiently discover trajectories that are certified to be physically feasible by the simulation.

We demonstrate our method on eight challenging scenarios where nonprehensile manipulation is required to successfully grasp the object. Our method is able to discover diverse strategies that successfully completes the grasping tasks while satisfying physical constraints. We compare our method to kinematic-based motion generation(~\cite{manipnet}) and show the motions generated by our method is visually superior with no hand-object interpenetration or telekinetic interaction.

\section{Related Work}
Manipulating objects with a dexterous hand is a long-standing research challenge that interests both the graphics and robotics communities. In this section, we review literature on physically plausible manipulation planning. 



\subsection{Dexterous manipulation without environmental interaction}
Due to the increased complexity when considering physical laws, many existing contributions assume either that manipulation tasks are done in free space, or that the hand-object interaction is the only relevant interaction. We review two of the most discussed manipulation tasks in this domain.


\subsubsection{Dexterous Grasping}
Grasping describes the task of generating hand and finger configurations to firmly holds an object of interest. To synthesize grasping motion analytically, some works leverage motion data to design control laws~\cite{pollard2005physically}, compute contact interactions~\cite{kry2006interaction}, and formulate optimization problems~\cite{zhao2013robust}. Other works assume the object trajectory is known and use it as a basis to synthesize hand motion through trajectory optimization~\cite{optim_manip, contactsampling,  animatemanip}. Additionally, physics-inspired grasp metrics, such as matching geometry~\cite{li2007data}, wrench closure and no collision \cite{closurecriteria, eigengrasp, graspevaluation}, are commonly applied in these works. More recently, advances in deep generative models(e.g., \cite{cvae,gan}) has given rise to learning-based grasp-generation methods for different object geometries~\cite{grasptta,grabnet,manohand, grasplearning1, grasplearning2, grasplearning3, grasplearning4}, \textcolor{black}{some of which also ensures physics feasibility and stability of the grasp \cite{wu2022learning,dgrasp}.}  The common limitation of these methods is that they only produce grasps for non-occluded object in free space. We note that there is another branch in the grasping literature, often dubbed as ``grasping in clutter'' or ``bin picking,'' which studies cluttered-scene object picking with simple grippers. As these works seldom use a dexterous hand, we excluded them from our review.

\subsubsection{In-hand manipulation}
In-hand manipulation seeks to move an object in a dexterous hand to a desired pose relative to the hand, using only the hand itself. In recent years, in-hand manipulation has become a popular task in the robotics community as a benchmark for challenging physical interaction. Mordtach and colleagues \cite{cio} formulate in-hand manipulation as a trajectory optimization problem with hand-object contact constraints. More recently, reinforcement learning using physics-based simulation has been applied to reorienting objects \cite{openai_inhand,general_inhand,qi2022hand} and solving a Rubik's cube \cite{rubik}. The complexity of in-hand manipulation originates from the hand-object interaction. Environment contacts do not need to be considered in this task.

\subsection{Extrinsic dexterity: dexterous manipulation without environmental interaction}
Nonprehensile manipulation with \textit{extrinsic dexterity} \cite{nonprehensile} aims to utilize external contact forces from environment. This is especially useful for manipulating objects in cluttered space and greatly expands the scope of possible actions. For instance, extrinsic dexterity may facilitate downstream tasks by exposing previously occluded parts of an object in clutter~\cite{pushtosee}. Common strategies that rely on external contact interactions include pushing, pivoting and tilting \cite{pushingpivoting, eppner_ijrr}. Among these strategies, pushing has been most extensively investigated. Researchers have explored how to manipulate objects on a two-dimensional plane with up to two active contact points applied from a robot \cite{mitcsg, 2dpushing1,2dpushing2,2dpushing3, pushtosee}. Pivoting has also been used in robotics for robot to grasp objects that are initially not graspable \cite{wenxuan,sun2020learning}. \textcolor{black}{Combining those primitives, \cite{eppner_planner} proposed a graph-based planner for a single DoF dexterous hand to grasp  objects. However, general non-prehensile manipulation requires more complex motion planning and often resorts to sampling-based planner \cite{parallelplanning}.} Nevertheless, all these works use either a fixed-geometry manipulator, simple parallel jaw grippers or underactuated multi-finger hand. \textcolor{black}{Motion synthesis for fully actuated dexterous hand is more challenging because both finger movement planning and grasp generation become much more complex.} To the best of our knowledge, no existing work is capable of generating physically plausible motion sequences for a fully actuated multi-fingered dexterous hand using extrinsic dexterity.

\section{Preliminary}
\label{sec:preliminary}
Before describing our method, we first formally define the ``feasibility'' of a grasp. A grasp is represented by at most five contact points on the object surface, corresponding to five finger tips: [$\bm{p}^1$,$\bm{p}^2$,$\bm{p}^3$,$\bm{p}^4$,$\bm{p}^5$]. A feasible grasp needs to be both \textit{dynamically} and \textit{kinematically} feasible~\cite{wu2022learning}.

\textit{Dynamic feasibility} requires each finger tip to apply a minimal normal force $f_\text{min}$ on the object while maintaining zero net wrench. This is referred to as \emph{wrench closure} \cite{graspit}. Meanwhile, the contact force applied on each contact point must be bounded within the friction cone specified by the friction coefficient $\mu$. To summarize, the two requirements for a dynamically feasible grasp are:

\begin{align*}   
\label{eq:qp}
&\min_{\bm{f}} \|\sum_{i=1}^5 \bm{f}^i\|_2^2 + \|\sum_{i=1}^5 \bm{p}^i \times \bm{f}^i\|_2^2 = 0, \\
 &\text{subject\;to}\;\; 0 < f_\text{min} < -\bm{f}^i \cdot\hat{\bm{n}}^i, \forall i\in\{1,\cdots,5\}\nonumber \\
 &|\bm{f}^i\cdot\hat{\bm{t}}^{i,j}| \leq \mu\bm{f}^i\cdot\hat{\bm{n}}^i, \forall i\in\{1,\cdots,5\}, \forall j\in\{1,2\} 
\end{align*}
where $\bm{f}^i$ is the contact force at the contact point $\bm{p}^i$ and $\hat{\bm{n}}^i$ is the surface normal at the contact point. $\hat{\bm{t}}^j,\forall j\in\{1,2\}$ are orthogonal basis vectors used to approximate the projection of the friction cone.  If the above constrained quadratic optimization problem can be solved with zero optimal value, the grasp is dynamically feasible. 

\textit{Kinematic feasibility} requires the contact points to be reachable by the dexterous hand. Such reachability can be confirmed by solving the inverse kinematics problem for the given contact points.

\section{Method}

We introduce a method to synthesize nonprehensile manipulation given the point cloud of the object of interest $\mathcal{O}$, the environment state relative to the initial object pose $\mathcal{S}_0$, the learned \emph{score function} $f_\theta: (\bm{p}^1, \bm{p}^2,\mathcal{S}, \mathcal{O}) \mapsto \mathbb{R}$, and the learned \emph{grasp generator} $g_\phi: (\bm{p}^1, \bm{p}^2, \mathcal{O}) \mapsto (\bm{p}^1,\bm{p}^2, \bm{p}^3, \bm{p}^4, \bm{p}^5)$. The initial environment state $\mathcal{S}_0$ is represented as a signed distance function (SDF) from the point cloud of the object to its surrounding. $\bm{p}^i, i=1\cdots5$ indicates the position of the five fingertips of an anthropomorphic hand, ordered from the thumb to the little finger and represented in the object coordinate frame. Given the object state, the thumb's and the index finger's locations on the object, the score function $f_\theta$ evaluates how likely such a two finger contacts may lead to a feasible grasp. With the same finger contact locations, the grasp generator $g_\phi$, which is the decoder of a conditional variational autoencoder (CVAE), predicts the best locations for the other three fingers. $f_\theta$ and $g_\phi$ are produced by an offline training process described in Section \ref{sec:grasp_gen_score_func}.

\begin{figure*}[t]
  \centering
  \includegraphics[width=0.8\linewidth]{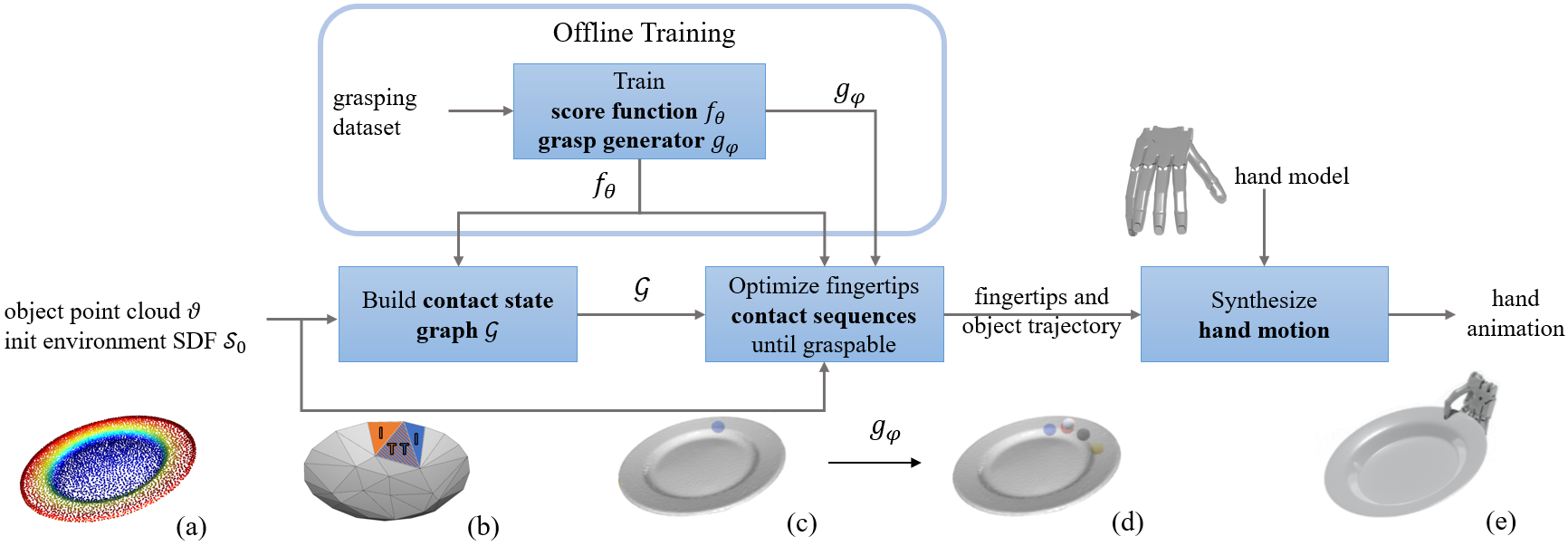}
  \caption{An overview of our pipeline. Using plate grasping as an example, (a)-(e) illustrate the steps. (a) Input point cloud. (b) Contact state graph. (c) Thumb and index fingertip motion, as well as object motion obtained from the trajectory optimizer. (d) A full grasp generated conditioned on the last frame of the trajectory. (e) Resulting hand motion after the IK process.}
  \label{fig:pipeline}
\end{figure*}

Our method consists of three steps: 1) Construct the contact state graph; 2) Optimize dynamically feasible contact trajectories and the object trajectory for nonprehensile manipulation; and 3) Synthesize animation. We first construct a contact state graph $\mathcal{G}$ based on the input object $\mathcal{O}$, the initial environment SDF $\mathcal{S}_0$, and the learned score function $f_\theta$. Since at most two finger contacts are required to achieve nonprehensile pregrasp, we can efficiently optimize contact trajectories of the thumb and the index finger on the contact state graph via a sampling-based optimization process. We take a learning approach to define optimality that rewards the hand to manipulate the object into a graspable configuration in a cluttered environment, subject to dynamic constraints. Finally, we solve a sequence of inverse kinematic (IK) problems to produce the animation of a dexterous hand, conditioned on the fingertip contacts and the object motion produced by the previous step. Figure \ref{fig:pipeline} gives an overview of our method.


\begin{algorithm}[b!]
\caption{Building contact state graph}
\KwIn{Object point cloud $\mathcal{O}$; SDF of environment obstacles $\mathcal{S}_0$}
$\mathcal{M}\longleftarrow\texttt{create\_mesh}(\mathcal{O})$\\
$\mathcal{V}_\text{all} \longleftarrow \texttt{ordered\_triangle\_pairs}(\mathcal{M})$\\
$\mathcal{S}_c \longleftarrow \texttt{calc\_nodal\_score\_in\_context}(\mathcal{V}_\text{all}, f_\theta)$\\
$\mathcal{S}_n \longleftarrow \texttt{calc\_nodal\_score\_no\_context}(\mathcal{V}_\text{all}, f_\theta)$\\
$\text{\# Remove low scoring nodes}$\\
$\mathcal{V}_c\longleftarrow\texttt{select\_top\_M}(\mathcal{S}_c,\mathcal{V}_\text{all})$\\
$\mathcal{V} \longleftarrow \mathcal{V}_c \cup \texttt{select\_top\_M}(\mathcal{S}_n, \mathcal{V}_\text{all})$\\
$\text{\# Connect nodes with edges}$\\
$\mathcal{E} \longleftarrow\emptyset$\\
\For {$u,v\text{ in }\mathcal{V}$} {
    \If{$\mathcal{T}^1_u = \mathcal{T}^1_v\text{ or }\mathcal{T}^2_u = \mathcal{T}^2_v$}{
        $\mathcal{E}\texttt{.add}(u,v)$
    }
}
$\mathcal{G} \longleftarrow \{\mathcal{V}, \mathcal{E}\}$\\
$\textbf{Output: }\text{Nodes for initial state }\mathcal{V}_c;\text{ Contact state graph }\mathcal{G}$\\
\label{alg:build_csg}
\end{algorithm}

\subsection{Contact state graph}
Given an object point cloud $\mathcal{O}$ and an initial environment SDF $\mathcal{S}_0$, we construct a contact state graph $\mathcal{G}$ to encode the relationships between different different regions of the object surface. $\mathcal{G}$ allows our method to efficiently explore the rich hand-object contact behaviors with optimization (Section \ref{subsec:traj_opt}). 

Algorithm \ref{alg:build_csg} summarizes the process of constructing the contact graph $\mathcal{G}$. We first approximate the object surface by fitting a mesh containing $30$ to $50$ triangles to $\mathcal{O}$ via an off-the-shelf mesh simplification algorithm \cite{quadric_decimation}. Since we only need to consider the thumb and the index finger thanks to external contacts, we define each node in $\mathcal{G}$ as $\bm{v} = (\mathcal{T}^1, \mathcal{T}^2)$. $\mathcal{T}^1$ and $\mathcal{T}^2$ are the triangles in the mesh that the thumb and the index finger are in contact with respectively. Because many of the triangle pairs are not ideal contact locations for nonprehensile pregrasp, we use the learned score function $f_\theta$ to prune $\mathcal{G}$ and only keep the high-scoring nodes. For each node $\bm{v}$, we compute a score $s_c$ that approximates the success likelihood of the node within the context of the environment, and a score $s_n$ that does not consider the environment:
\begin{equation}
s_c = f_\theta(\bm{p}^1, \bm{p}^2, \mathcal{S}_0, \mathcal{O}'); \;\;\;\; s_n = f_\theta(\bm{p}^1, \bm{p}^2, \emptyset, \mathcal{O}),
\end{equation}
where $\mathcal{O}'$ is a subset of $\mathcal{O}$ with occluded points removed. 

We sample five fingertip locations within $\mathcal{T}^1$ and $\mathcal{T}^2$ associated with $\bm{v}$ for each of the three scenarios: thumb-only contact, index-finger-only contact, and two-finger contact. We query $f_\theta$ for these 15 fingertip placements and compute the average $s_c$ and $s_n$ for $\bm{v}$. If either $s_c$ or $s_n$ is within the top $M$ percentile among all nodes, we include $\bm{v}$ in $\mathcal{G}$. If $v$ is selected due to a high $s_c$, we put it in a subset of selected nodes called $\mathcal{V}_c$. This subset of nodes will be used for the initial contact state in the optimization process (Section \ref{subsec:traj_opt}). 

\begin{figure}[t]
  \centering
  \includegraphics[width=0.75\linewidth]{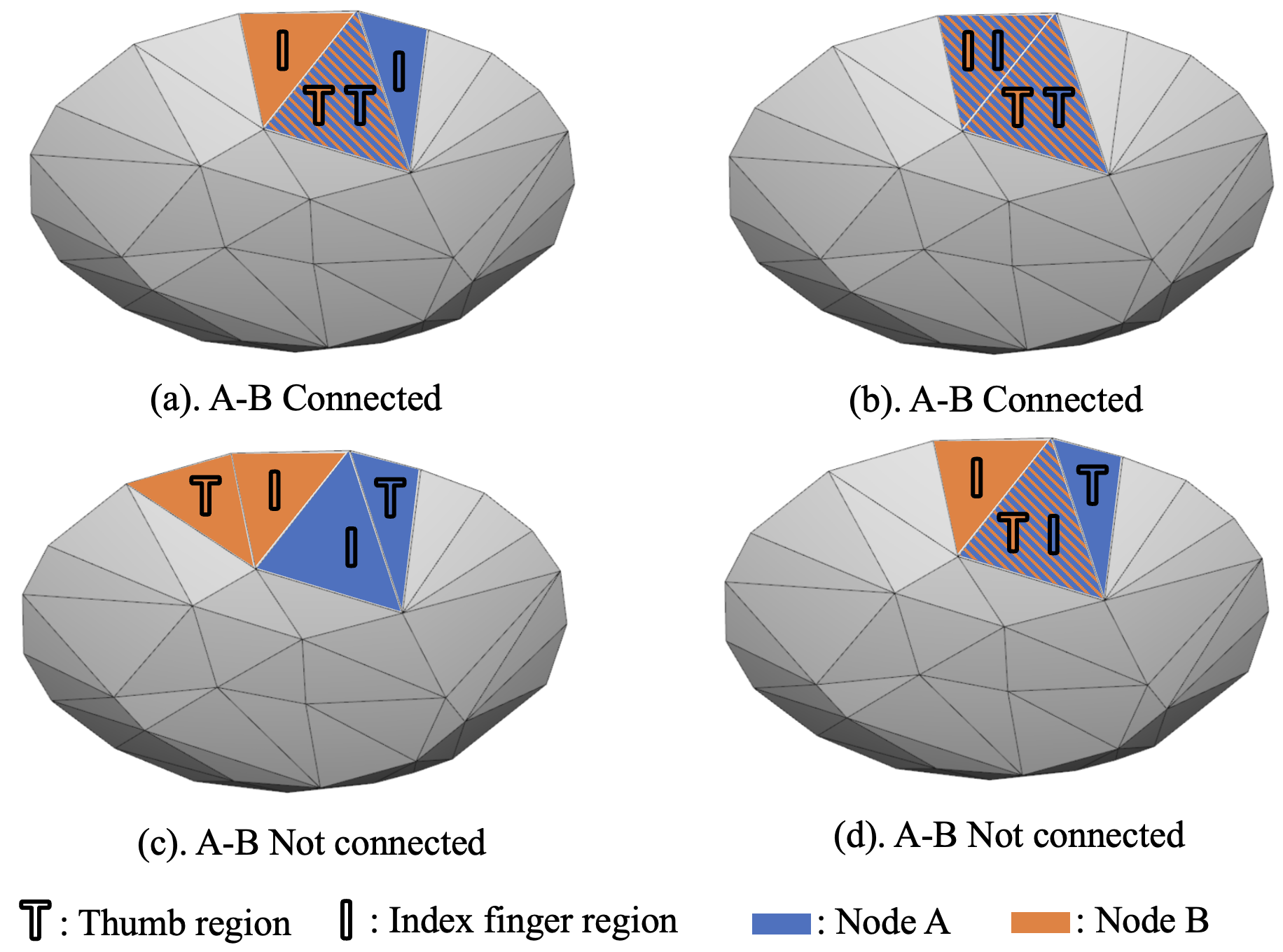}
  \caption{Different types of connectivity between nodes in the contact graph.}
  \label{fig:connectivity}
\end{figure}

Each edge of $\mathcal{G}$ represent a feasible contact state transition. To prevent both fingers from simultaneously changing contact locations, we only connect two nodes by an edge if they share either $\mathcal{T}^1$ or $\mathcal{T}^2$. Each node also connects to itself to allow for consecutive unchanged contact states. This is illustrated in Figure \ref{fig:connectivity}.

\subsection{Trajectory optimization}
\label{subsec:traj_opt}
Once $\mathcal{G}$ is constructed, the next step is to search for optimal trajectories for the thumb contact, the index-finger contact, and the 6D pose of the object such that a wrench-closure grasp using all five fingers can be achieved successfully at the end. The optimization is a double-loop iterative process. The outer loop solves a discrete path planning problem while the inner loop solves a continuous optimal control problem. Algorithm \ref{alg:traj_opt} shows the detailed procedure of this optimization.

A \emph{path} is a sequence of $N$ nodes connected by edges on $\mathcal{G}$. It determines the $N$ \emph{contact stages} of the animation. Since the object is positioned in the context of the environment initially, we restrict the first node of each path to be a node in the subset $\mathcal{V}_c$. The remaining $N-1$ nodes can be any node on $\mathcal{G}$. The score of a path is computed by accumulating the score at each contact stage:
\begin{equation}
\label{eqn:path_score}
s_{p} = s_c(\bm{v}_1) + \sum_{i=2}^{N} s_n(\bm{v}_i),
\end{equation}
where $s_c$ and $s_n$ are overloaded to indicate functions that return the respective score of the input node. Using Equation \ref{eqn:path_score}, we compute the scores for all possible paths and sort the paths by their scores from high to low.

The outer loop of the optimization iterates over the sorted list of paths until a successful trajectory is found by the inner loop. For each given path $(\bm{v}_1, \cdots, \bm{v}_{N})$, the inner loop solves for contact locations of the fingertips subject to staying within the triangles associated with the nodes in the path. Formally, we optimize the state variable $\bm{x}_i = (\bm{p}^1_i, \bm{p}^2_i, b^1_i, b^2_i)$ where $i = 1 \cdots N$, and the control variables $\bm{u}_t = (\bar{\bm{p}}^1_t, \bar{\bm{p}}^2_t)$ where $t=K \cdot N$. $\bm{p}^1_i$ and $\bm{p}^2_i$ are the thumb position and the index-finger position in the object coordinate frame at contact stage $i$. To distinguish between thumb-only, index-finger-only, or two-finger contact scenarios, we define binary variables $b^1_i$ and $b^2_i$ to indicate whether the respective finger is in contact with the object at contact stage $i$. We use a proportional derivative-like control scheme to control the contact forces exerted by the fingers on the object: $\bm{f} = -k (\bm{p}^j - \bar{\bm{p}}^j)$, where $k=50$ is a predefined coefficient. $\bm{p}^j$ and $\bar{\bm{p}}^j$ are the current fingertip location and its target location. We define the control variables $(\bar{\bm{p}}^1_t, \bar{\bm{p}}^2_t)$ as the target locations of the thumb and the index fingertip in the world coordinate frame at frame $t$. Because the contact forces tend to change at a higher rate than the contact locations, we allow the control variables to change $K$ times within each contact stage. 

The objective function of the inner optimization is as follows:
\begin{equation}
\mathcal{L}(\bm{x}_{1:N}, \bm{u}_{1:KN}) = \sum_{i=i}^N\; f_\theta(\bm{p}^1_i \wedge b^1_i, \bm{p}^2_i \wedge b^2_i, \mathcal{S}_i, \mathcal{O}'_i), 
\end{equation}
where we use $\bm{p}\wedge b$ to denote ``$\bm{p}$ if $b$ is true, $\emptyset$ otherwise.'' Note that the environment SDF $\mathcal{S}_i$ depends on the object pose and is essentially a function of $\bm{x}$ and $\bm{u}$ via physics simulation. We compute $\mathcal{S}_i$ as the environment SDF corresponding to the object pose at the beginning of contact stage $i$. Similarly, we remove occluded points from $\mathcal{O}$ to form $\mathcal{O}'_i$ based on the object pose at the beginning of contact stage $i$.

To ensure the contact locations staying within the boundary of the given triangles, we use barycentric coordinates to parameterize $\bm{p}^1$ and $\bm{p}^2$. To further ensure the physical plausibility of the finger motion, a finger can only change its location after it is detached from the object in the previous contact stage. For example, $b^2_{i-1}$ must be false for the index finger to change the location at contact stage $i$. This additional constraint prevents fingers from jumping instantaneously from one location to another.

We use a sampling-based optimizer, MPPI \cite{williams2017model} since gradient-based optimization algorithms tend to be stuck in local minima when solving contact-rich control tasks \cite{bundlegradient}. After solving each trajectory optimization problem, we query the grasp generator $g_\phi$ to complete a five-finger grasp based on the final locations of thumb and/or the index fingertips, as well as the object point cloud:
\begin{equation}
    (\bm{p}^1, \bm{p}^2, \bm{p}^3, \bm{p}^4, \bm{p}^5) = g_\phi(\bm{p}^1_N, \bm{p}^2_N, \mathcal{O})
\end{equation}
There are three possible outcomes during the nonprehensile manipulation phase: 1) only thumb is in contact, 2) only index finger is in contact 3) both thumb and index finger are in contact. [100,100,100] is used as a invalid token if a finger is not in contact.
Lastly, we test whether a force-closure grasp can be formed by any combination of the three finger contacts from the set ($\bm{p}^1_N$, $\bm{p}^2_N$, $\bm{p}^3$, $\bm{p}^4$, $\bm{p}^5$). Since $g_\phi$ is the decoder of a CVAE, we sample the latent space $20$ times to generate $20$ different grasps. If any one of them is feasible, we exit the outer loop and the optimization is complete.

\begin{algorithm}[b!]
\caption{Pregrasp nonprehensile manipulation optimization}
\KwIn{Contact state graph $\mathcal{G}$; Initial nodes $\mathcal{V}_c$; Path length $N$}
$\mathcal{P}\longleftarrow \texttt{all\_paths\_from\_graph}(\mathcal{V}_c, \mathcal{G}, N)$\\
$\text{\# Compute path score}$\\
$\mathcal{S}_\mathcal{P}\longleftarrow\emptyset$\\
\For{$\bm{V}_\text{path}\text{ in }\mathcal{P}$}{
    $s_p\longleftarrow s_c(v_1)+\sum_{i=2}^N{s_n(v_i)}$\\
    $\mathcal{S}_\mathcal{P}\texttt{.add}(s_p)$\\
}
$\mathcal{P}_\text{sorted} \longleftarrow \texttt{sort\_by\_score}(\mathcal{P}, \mathcal{S}_{\mathcal{P}})$\\
$\text{\# Solve Trajectory Optimization}$\\
\For {$\bm{V}_\text{path} \text{in} \mathcal{P}_\text{sorted}$} {
    $\text{\# Do MPPI trajectory optimization}$\\
    $(\bm{x}^*_{1:N}, \bm{u}^*_{1:KN}) \longleftarrow \texttt{MPPI\_optimization}(\bm{V}_\text{path}, f_\theta)$ \\
    $\text{\# Generate grasp with grasp generator}$\\
    $\mathcal{O}, (\bm{p}^1_N, \bm{p}^2_N)\longleftarrow \texttt{simulate}(\bm{x}^*_{1:N}, \bm{u}^*_{1:KN})$\\   
    $\bm{p}^{1:5} \longleftarrow g_\phi(\bm{p}^1_N, \bm{p}^2_N, \mathcal{O})$ \\
    \If {$\bm{p}^{1:5}\text{ is feasible}$} {
        $\textbf{break}$\\
    }
}
\label{alg:traj_opt}
\end{algorithm}

\subsection{Grasp generator and score function}
\label{sec:grasp_gen_score_func}
\begin{figure}[t]
  \centering
  \includegraphics[width=0.8\linewidth]{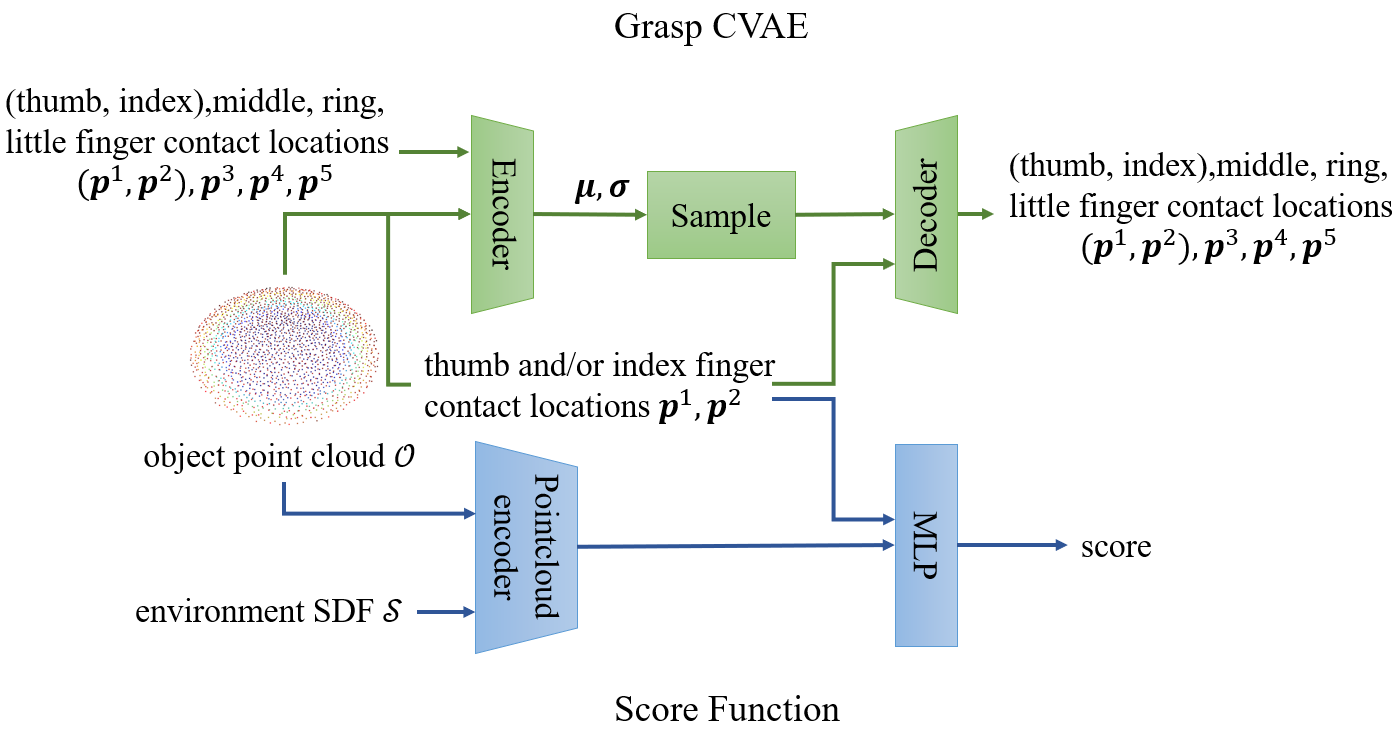}
  \caption{Overview of the grasp CVAE and the score function.}
  \label{fig:architecture}
\end{figure}

\paragraph{Grasp generator:}
We train the grasp generator, $g_\phi$, as three conditional variational autoencoders (CVAE). Once trained, the decoder is used to predict the contact locations for the middle, ring, and little fingers conditioned on the initial contact locations of the thumb and/or the index finger, as well as the object point cloud expressed in the object coordinate frame (Figure \ref{fig:architecture}). The training data for the grasp generator is a synthetic dataset of feasible grasps of various objects in different environmental contexts. We create the dataset using the YCB object set \cite{ycb}. For each object, we first manually define 3 feasible grasps as seed grasps. For each seed grasp, we perturb the finger contact locations and check whether the perturbed grasp is still both kinematically and dynamically feasible. If so, the perturbed grasp becomes a new seed grasp and added to the dataset. The process repeats until sufficient feasible grasps are generated for this object. 

\paragraph{Score function:}
Given the thumb and the index finger contact locations, the environment SDF, and the object point cloud expressed in the object coordinate frame, we train a Multilayer Perceptron (MLP), $f_\theta$, to evaluate how likely the two finger contacts will lead to a successful feasible grasp (Figure \ref{fig:architecture}). We use PointNet++ \cite{pointnet++} to encode the input point cloud and the SDF. The output encoding is concatenated with the contact locations of the thumb and the index finger to form the input of the MLP. We leverage the grasp generator to create the labels for our training data. For each data point, we first randomly select one or two contact points on the object which are allocated as thumb and/or index finger contact location. Conditioned on these contact points, we sample the latent space of the corresponding grasp generator $P$ times to complete $P$ grasps. We then check the feasibility of the $P$ grasps and assign the label as ``the ratio of the number of feasible grasps to $P$.''


\subsection{Hand motion synthesis}
\label{subsec:ik}
Given the finger contact trajectories and the object trajectory from the previous step, we solve an inverse kinematics (IK) problem to obtain the detailed hand pose for each frame of the final animation. We frame the IK problem as an optimization with an objective function that (1) ensures the fingertips of the hand satisfy the contact trajectories, (2) avoids collisions with the other objects in the environment and self-collision with the hand, and (3) encourages smoothness in the kinematic trajectory.

The solution to such a nonconvex optimization is very sensitive to the initial guess. A typical strategy is to solve the IK problem frame-by-frame in chronological order, so the decision variables of the current frame can be initialized with the solution from the previous frame for increaseed motion continuity. However, we found that using a reverse chronological order, multi-layered strategy to solve the animation frames results in higher quality animation. We first solve IK for the $N+1$ “keyframes”, which include the first frame of each contact stage and the last frame of the animation. Once all keyframes are solved, we can interpolate the hand poses to create initial guesses for the in-between frames. The key frames are solved in a reverse chronological order such that the current keyframe's optimization is initialized with the solution of the optimizating for the next keyframe in the optimization. This is because the final keyframe involves all five fingers, resulting in a more constrained IK problem than keyframes earlier in the animation sequence. 
In practice, the IK almost always solved. If the IK solver fails after multiple attempts, we make a slight compromise on quality and manually loosen the optimization's constraint tolerances, such as those for avoiding collision and matching desired contact locations.


\section{Evaluation}
We evaluate our method qualitatively by demonstrating a rich and diverse set of synthesized hand animations. We also provide quantitative analysis on our method's robustness, generalizability, and ablations. Finally, we compare with the most relevant method in literature, ManipNet\cite{manipnet}. We use a five-finger Shadow Hand model with $28$ degrees of freedom ($22$ joints and $6D$ wrist pose) in our implementation. 
Our approach can be applied to other hand models as long as all relevant training is done with the new hand model.
We use PyBullet \cite{pybullet} for physics simulation and MPPI \cite{williams2017model} for trajectory optimization. Inverse kinematics problems are solved using the collision-free IK Module in Drake \cite{drake}, which uses \cite{SNOPT} as the backend optimizer.

\subsection{Motion quality and diversity}
We design eight diverse scenarios to test our method, including (1) adjacent books on a bookshelf, (2) a ruler lying flat on the edge of a table, (3) a computer keyboard and (4) a flat cardboard placed on a shelf against a wall, (5) a dinner plate and (6) a food container placed on a the table, (7) a densely packed spice rack, and (8) a marker stored in a pencil box. We show that in all scenarios, our method is able to synthesize natural and physically plausible nonprehensile pregrasps to manipulate the objects into a graspable configuration. We observe four distinct strategies used by the dexterous hand: (1) repositioning the object to expose the bottom surface (ruler, cardboard), (2) pivoting the object against its surroundings (marker, keyboard, book, spice bottle), (3) utilizing finger friction (food container), and (4) utilizing geometric features (plate). Figure \ref{fig:tasks} and the supplemental video show animations of our results.

\begin{figure*}[h]
    \centering
    \includegraphics[width=0.8\linewidth]{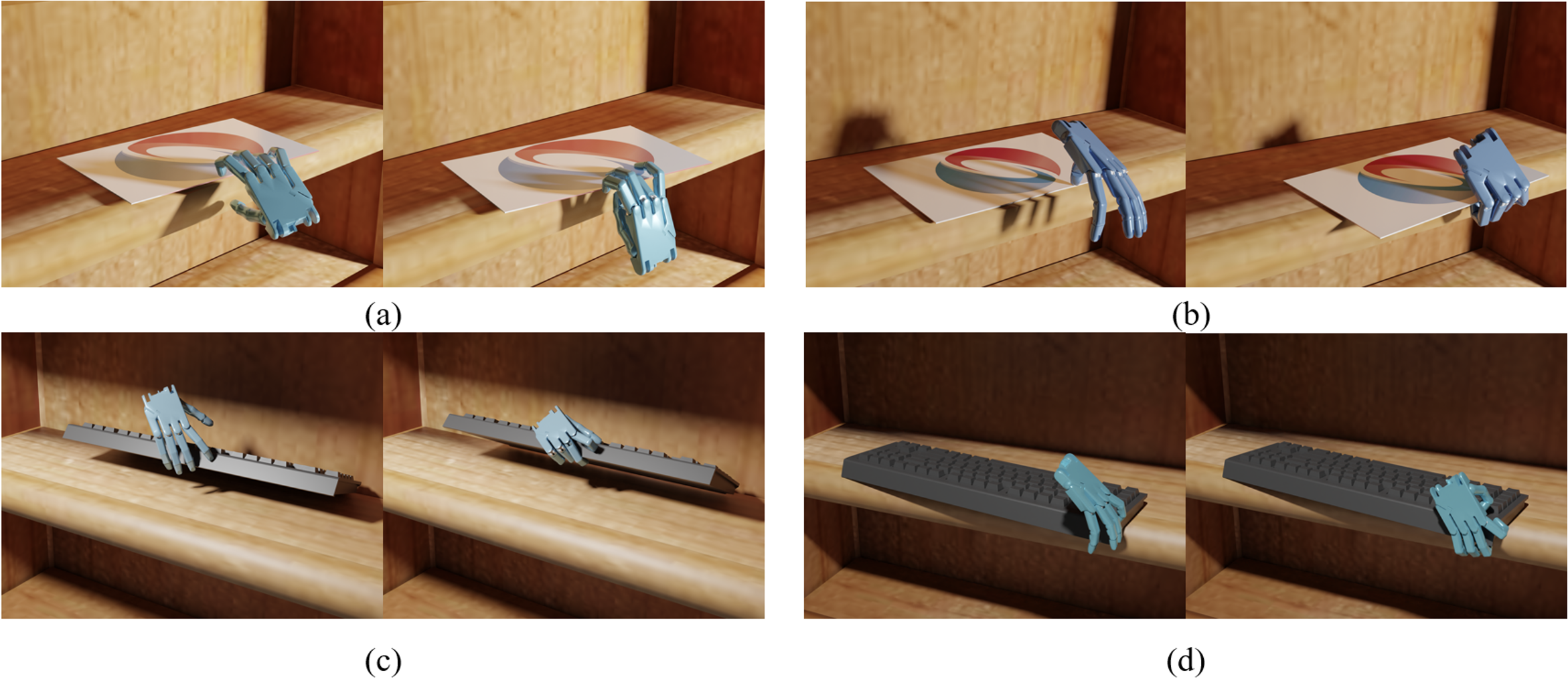}
    \caption{(a), (b) are of same configuration. (c), (d) are different. Keyboard in (c) is closer to the wall, in (d) it is closer to the edge.}
    \label{fig:diverse}
\end{figure*}

To further demonstrate the diversity of motions generated by our method, we showcase some scenarios where multiple successful pregrasp strategies exist. By searching over more paths on the contact graph instead of exiting upon the first successful path as in \ref{alg:traj_opt}, our method is able to generate visually and functionally different manipulation strategies. Figure \ref{fig:diverse}(a) and \ref{fig:diverse}(b) show two different ways to pick up the cardboard: one uses the index finger to pull the cardboard outward, while the other one uses the thumb. Changing the surrounding of the object can also result in different strategies. If the keyboard is placed near the backboard of the shelf, the hand will push the keyboard against the backboard to pivot it up (Figure \ref{fig:diverse}(c)). If the keyboard is placed near the edge of the shelf, the hand will lift the keyboard from the corner that is fully exposed.

\subsection{Quantitative evaluation of algorithms}
We define two metrics to evaluate our method. Based on these two metrics, we analyze the generalizability of our method to unseen objects and conduct ablation studies on the design of score function.

\textbf{Grasp success rate} Feasibility of the grasp generated by the grasp generator is the most critical indicator of the success of our method. A feasible grasp must satisfy both dynamic and kinematic constraints (Section \ref{sec:preliminary}). Since we use a sampling-based grasp generator (i.e. a CVAE), we compute the grasping success rate $r_\text{suc}$ based on the first feasible path. $r_\text{suc}$ is defined as the ratio between the number of feasible grasps and the total number of samples ($20$ in our implementation) drawn from the latent space.

\textbf{Number of paths attempted:} The sampling-based optimizer may need to sample multiple candidate contact paths before a feasible one is found. The number of paths attempted is another metric to measure the efficiency of our algorithm. For each candidate path, we test a set of $20$ grasps generated by different latent space samples of the grasp generator. If there exists one feasible grasp, the path is considered successful. The total number of paths attempted to obtain a feasible grasp is denoted as $N_\text{paths}$.

\begin{figure*}[h]
  \centering
  \includegraphics[width=\linewidth]{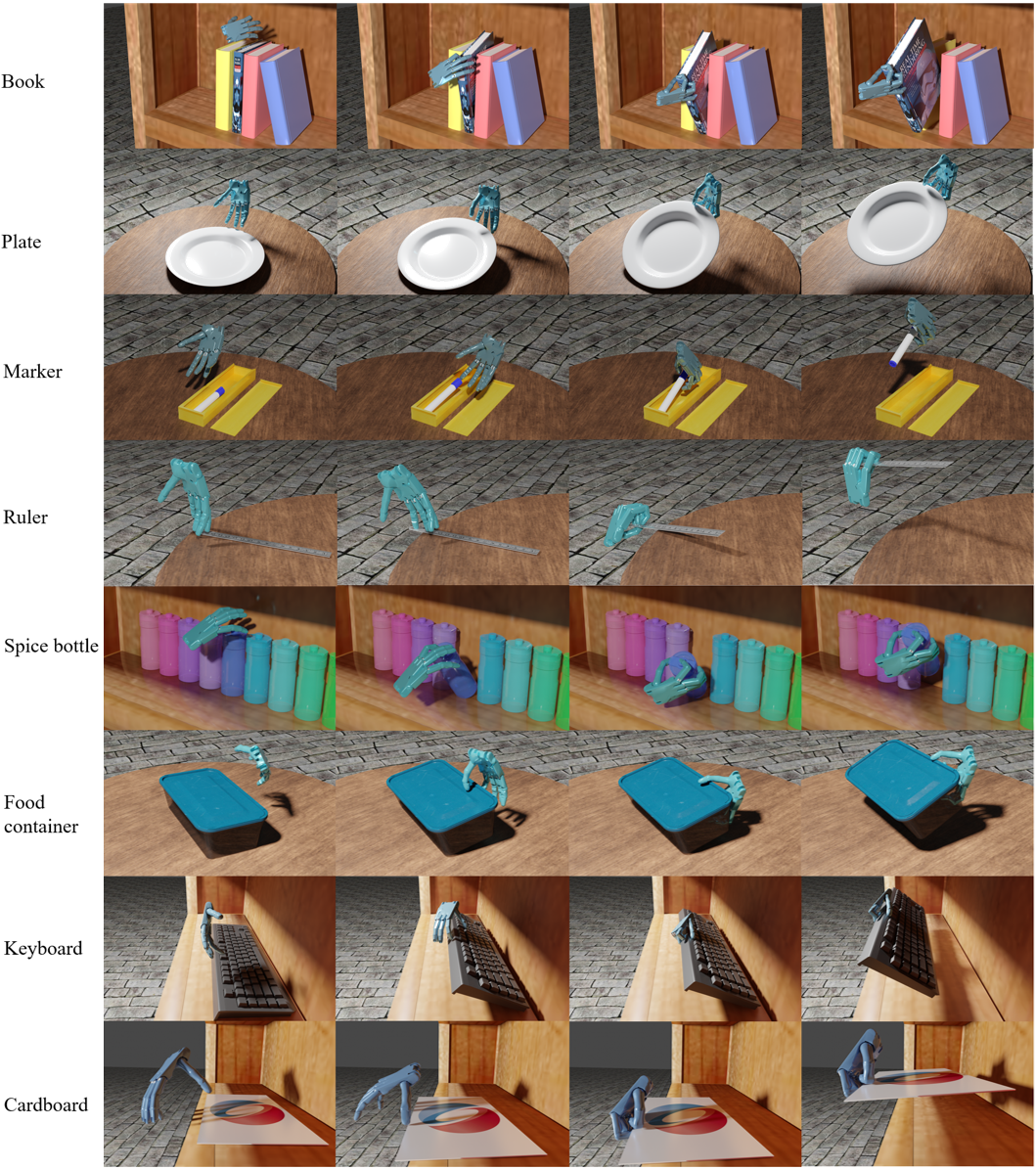}
  \caption{Results of our method in different scenarios}
  \label{fig:tasks}
\end{figure*}

We evaluate $r_\text{suc}$ and $N_\text{paths}$ on all eight scenarios in Figure \ref{fig:tasks} and report the results in Table~\ref{tab:quant_results}. The results demonstrate that our grasp generator can reasonably predict full grasp condition on final object pose and thumb and index finger contact location with high success rate. All tasks are completed within 20 minutes of runtime on a desktop computer with Intel i9-9900K and NVIDIA RTX2080. 

\begin{table}[h]
    \centering
    \resizebox{0.9\linewidth}{!}{
    \begin{tabular}{@{}lrrrr@{}}
    \toprule
    \multicolumn{5}{c}{$r_\text{suc}$ and $N_\text{path}$ of all tasks}\\
    \hline
    &Bookshelf & Plate & Marker & Ruler\\
    \midrule
    $r_\text{suc}\ \ \ \ $     & 0.917(0.058) & 0.800(0.050) & 0.933(0.058) & 0.817(0.144)\\
    \hline
    $N_\text{path}\ \ \ \ $    & 1.667(1.155) & 1.667(1.155) & 3.667(2.082) & 3.333(1.528)\\
    \hline
    \hline
    \toprule
    & Waterbottle & Food container  & Keyboard & Cardboard\\
    \midrule
    $r_\text{suc}\ \ \ \ $     & 0.850(0.180) & 0.733(0.340) & 0.817(0.076) & 0.800(0.229)\\
    \hline
    $N_\text{path}\ \ \ \ $    & 3.000(1.732) &  2.000(1.000) & 1.667(0.577) & 1.333(0.577)\\
    \bottomrule
    \end{tabular}
    }
    \caption{Quantitative result of all environments. Means and standard deviations are obtained from 3 experiments. Computational budget is 10 paths.}
    \label{tab:quant_results}
\end{table}

\subsection{Generalization to unseen objects}
\begin{figure}[h]
  \centering
  \includegraphics[width=0.7\linewidth]{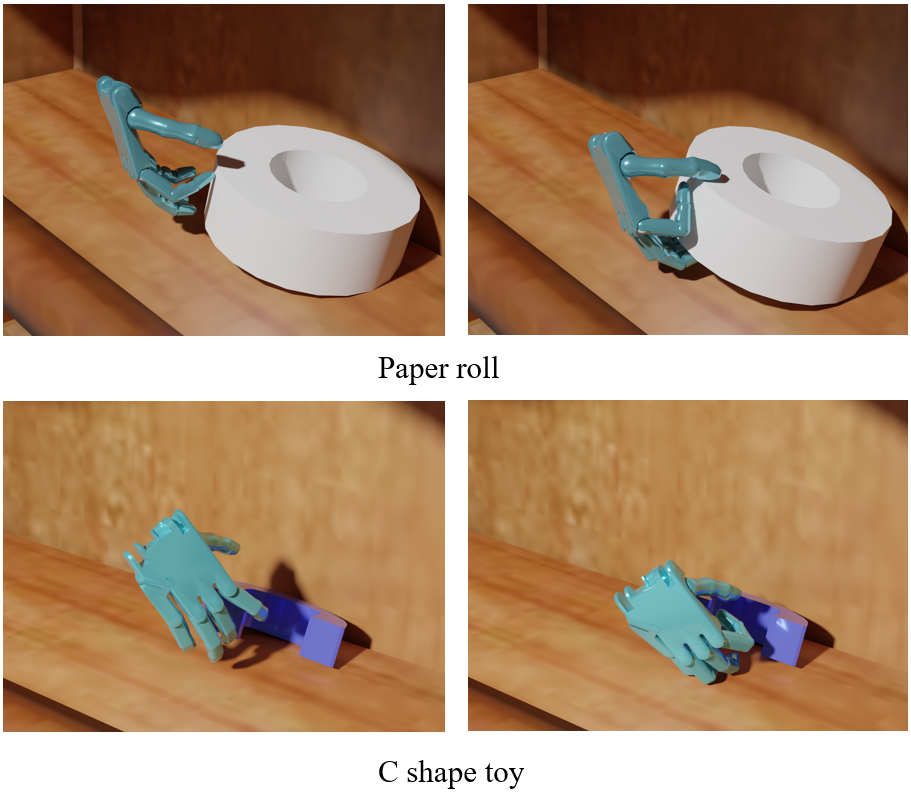}
  \caption{Performance of our method on nonconvex objects.}
  \label{fig:non_convex}
\end{figure}

Since our method contains learning-based components trained on a dataset of objects, it is crucial to demonstrate generalizability to unseen objects. All eight aforementioned scenarios were tested with unseen objects not included in the YCB training dataset. In addition, we evaluate the method on two nonconvex objects. For those objects that can be reasonably approximated by their convex hulls, such as a paper roll and a cookie jar, our method can successfully synthesize physically plausible motions. However, for highly non-convex objects, our current implementation is limited by two components that requires convexity: the mesh reconstruction algorithm, and the collision avoidance objective in Drake's IK solver (Figure \ref{fig:non_convex}). More failures and limitations are in our video.

We also evaluate generalizability across object size scaling using $r_\text{suc}$ and $N_\text{paths}$ as metrics. Table \ref{tab:scale} shows that our method could perform reasonably well with different scale of object in Keyboard environment. In more challenging Bookshelf environment, changing object scale may affect object's physics property such as center of mass and accessibility, causing performance fluctuation.

\begin{table}[h]
    \centering
    \resizebox{0.9\linewidth}{!}{
    \begin{tabular}{@{}lrrrr@{}}
    \toprule
    \multicolumn{5}{c}{$r_\text{suc}$ and $N_\text{path}$ of different scales}\\
    \hline
    &\multicolumn{2}{c}{Bookshelf} & \multicolumn{2}{c}{keyboard}\\
    \midrule
    scale(x,y,z)& $r_\text{suc}$   & $N_\text{path}$ & $r_\text{suc}$  & $N_\text{path}$ \\
    \hline
    [1,1,1]    & 0.917(0.058) & 1.667(1.155) & 0.816(0.076) & 1.667(0.577)\\
    \hline
    [0.5,1,1]  & 0.967(0.029) &  1.000(0.000) & 0.683(0.126) & 1.333(0.577)\\
    \hline
    [1,0.5,1]  & $\text{NA}^*$ &   $\text{NA}^*$ & 0.816(0.104) & 2.333(1.155)\\
    \hline
    [1,1,0.5] & 0.833(0.104) & 1.667(0.577) & 0.900(0.050) & 2.333(1.528)\\
    \hline
    [1,1,0.02] & $0.500(0.071)^*$ & $2.000(1.414)^*$ & 0.800(0.218) & 3.000(2.000) \\ 
    \bottomrule
    \end{tabular}
    }
    \caption{Results of various object sizes. Means and standard deviations are obtained from 3 optimizations. Computation budget is 10 paths. * indicates the algorithm occasionally failed within budget. $\text{NA}^*$ means all 3 executions failed.}
    \label{tab:scale}
\end{table}

\subsection{Ablation on score function design}

We perform an ablation study on the design choice of the score function. We evaluate two variants of the score function: 1) No SDF: only use positions of the points on the object surface as input to the point cloud encoder; 2) No Shape: only use signed distance values as input to the point cloud encoder. We compute $r_\text{suc}$ and $N_\text{paths}$ on these two variants using the plate and the marker examples. \ref{tab:ablation} shows that having both SDF and object shape information is crucial for accurately assessing the score of a grasp.

\begin{table}[h]
    \centering
    \resizebox{0.9\linewidth}{!}{
    \begin{tabular}{@{}lrrrr@{}}
    \toprule
    \multicolumn{5}{c}{$r_\text{suc}$ and $N_\text{path}$ of score function variants}\\
    \hline
    &\multicolumn{2}{c}{Plate} & \multicolumn{2}{c}{Food container}\\
    \midrule
    & $r_\text{suc}$   & $N_\text{path}$ & $r_\text{suc}$  & $N_\text{path}$ \\
    \hline
    Ours     & 0.800(0.050) & 1.667(1.155) & 0.733(0.340) & 2.000(1.000)\\
    \hline
    No SDF  & 0.583(0.058) &  5.333(2.887) & 0.450(0.436) & 2.667(1.528)\\
    \hline
    No pretrain  & 0.400(0.132) &  6.333(1.155) & NA & NA\\
    \bottomrule
    \end{tabular}
    }
    \caption{Result of the ablation study. Means and standard deviations are obtained from 3 experiments. Computational budget is 10 paths. NA denotes infeasible with budget.}
    \label{tab:ablation}
\end{table}

\subsection{\textcolor{black}{Ablation on different orders of solving IK}}
\begin{figure*}[h]
  \centering
  \includegraphics[width=0.8\linewidth]{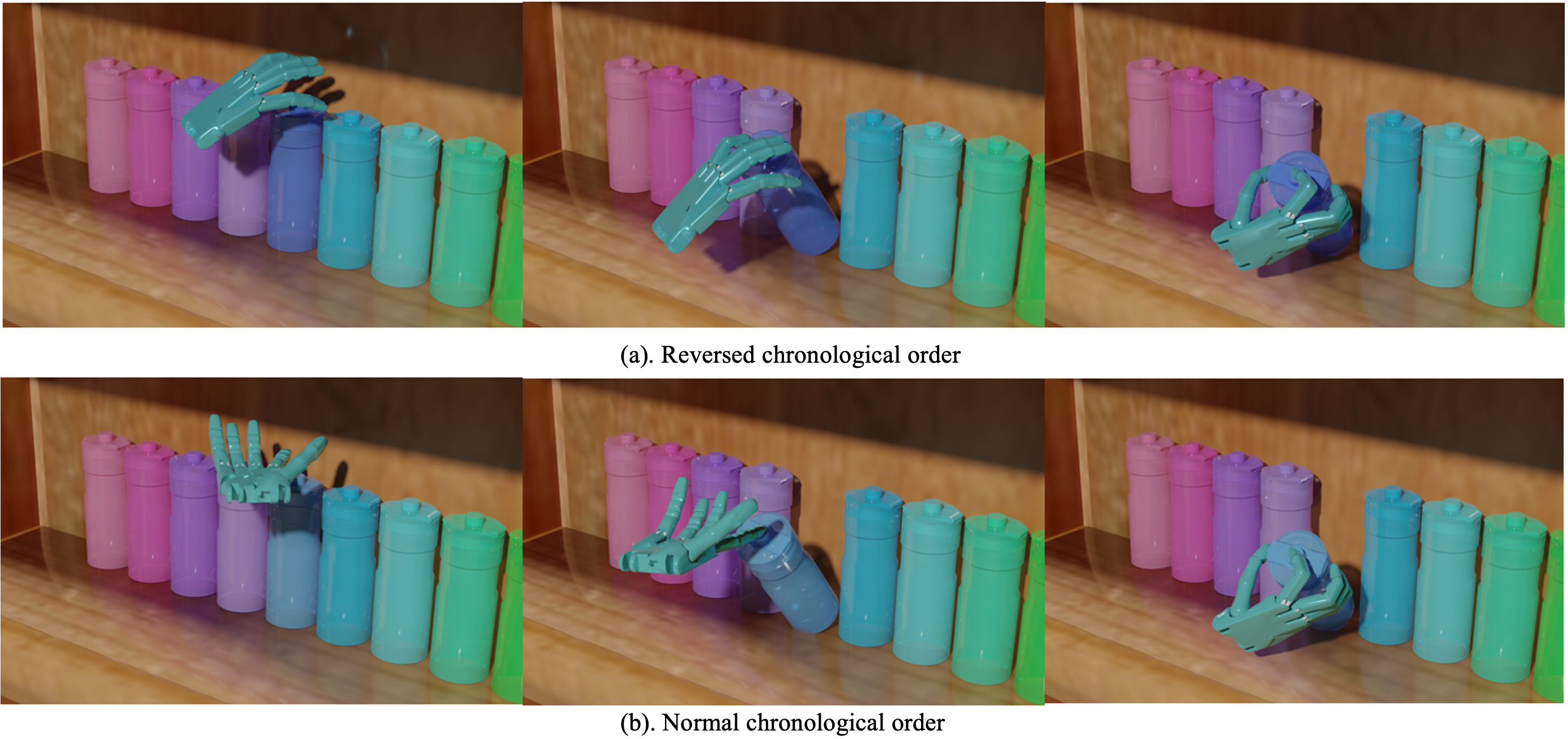}
  \caption{Comparison of normal and reverse order of solving IK.}
  \label{fig:order}
\end{figure*}

\textcolor{black}{We compare solving sequential IK in a forward  and reverse order. It shows that the reverse order achieves better consistency (Figure \ref{fig:order}). Solving IK in a forward order results in palm flipping abruptly because the IK solver cannot foresee the next grasping pose.} 

\subsection{Comparison with ManipNet}
\begin{figure}[h]
  \centering
  \includegraphics[width=0.8\linewidth]{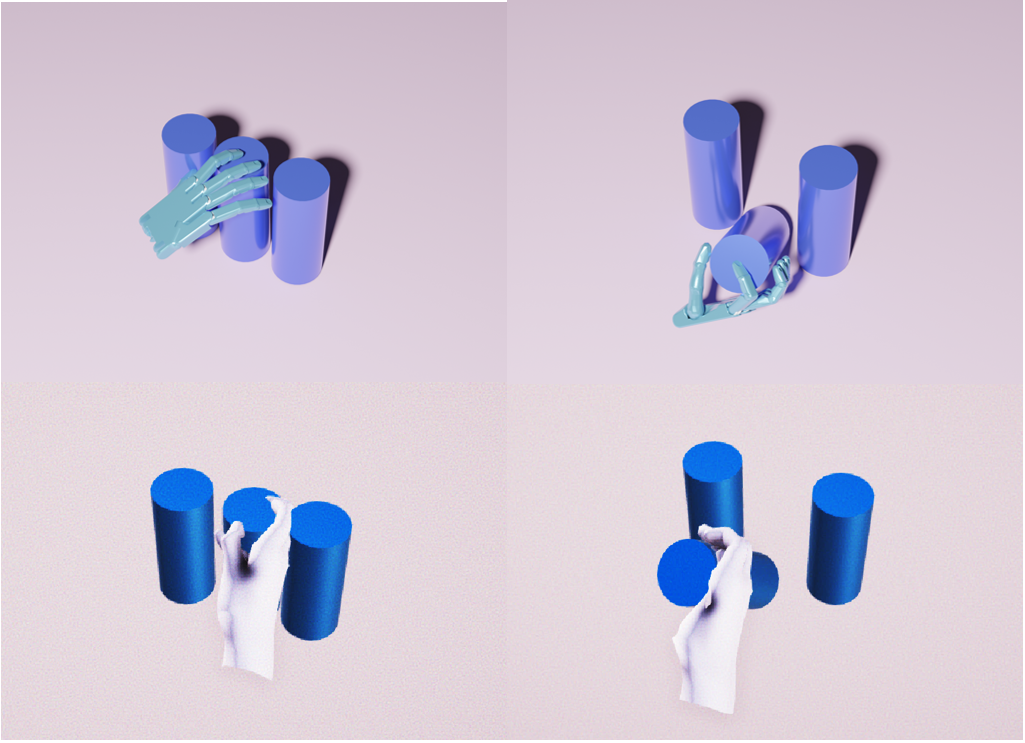}
  \caption{Comparison with ManipNet on spice rack.}
  \label{fig:compare_manipnet}
\end{figure}

Due to the lack of literature on non-prehensile manipulation for dexterous hand, we compare our method with the state-of-the-art, kinematics-based method, ManipNet \cite{manipnet}, in a scenario similar to the spice rack example. We pick this example because ManipNet is trained on grasping objects with similar cylinder shapes. In experiment, the object and wrist trajectory is obtained from running our method as ManipNet assumes this information as input. Figure \ref{fig:compare_manipnet} shows that, despite the ability to sense the manipulated object, ManipNet attempts to grasp the bottle without considering the surrounding objects. This results in significant finger-object interpenetration. In contrast, our method pulls out then grasps the bottle while avoiding collision with other objects. While the comparison may arguably be more fair if ManipNet was trained with similar context, recording training data that covers all possible surroundings for all different objects is impractical. This makes ManipNet difficult to extend to contextual environments.

\section{Conclusions}
We propose a physics-based method for synthesizing nonprehensile pregrasp animations that grasp an initially ungraspable object. 
Our method leverages extrinsic dexterity and uses a learned function to evaluate the ``graspability'' of an object in the context of the environment. We show that our method is capable of discovering a diverse set of pregrasp strategies and producing realistic and physically plausible hand and object motions.

\section{Limitations}
Our method has a number of limitations. First, the duration of contact stages is predefined and may lead to unnatural behaviors in some scenarios. Nevertheless, optimizing the timing of pregrasp behaviors is possible once motion data is available.  Our approach is also limited to grasping rigid object as the contact graph construction assumes the distances between triangles on the mesh are fixed. \textcolor{black}{Furthermore, the point contact assumption in this work makes two-finger grasps unstable and challenging to achieve.} A potential extension is to incorporate other contact points, such as the palm and the knuckles, for exploring different grasp strategies. Lastly, our pipeline is limited by a number of implementation choices. Currently, it cannot manipulate nonconvex objects that are poorly approximated by their convex hulls due to the selected mesh reconstruction method and IK solver. Improving the implementation can remove the convexity requirement and speed up the pipeline.
\begin{acks}

We'd like to thank Yifeng Jiang for designing action space.
The work is supported by the NSF:CCRI:2120095, Toyota Research Institute(TRI) and Stanford Institute for Human-Centered AI(HAI).

\end{acks}
\newpage

\bibliographystyle{ACM-Reference-Format}
\bibliography{reference}
\end{document}